\documentclass[5p,times]{elsarticle}

\usepackage{amssymb}
\usepackage{glossaries}

\usepackage{graphicx}
\usepackage{adjustbox}
\usepackage[usestackEOL]{stackengine}
\usepackage{multirow}
\usepackage{longtable, tabu}
\usepackage{tabularray}
\usepackage{lscape}
\usepackage{tabularx}
\usepackage{afterpage}
\usepackage{color,soul}
\usepackage{multicol}

\usepackage[table]{xcolor}

\definecolor{whitesmoke}{rgb}{0.96, 0.96, 0.96}
\definecolor{floralwhite}{rgb}{1.0, 0.98, 0.94}

\newacronym{IoT}{IoT}{Internet of Things}
\newacronym{DT}{DT}{Digital Twin}
\newacronym{PT}{PT}{Physical Twin}
\newacronym{DS}{DS}{Digital Shadow}
\newacronym{CDT}{CDT}{Cognitive Digital Twin}
\newacronym{SSN}{SSN}{Semantic Sensor Network}
\newacronym{SOSA}{SOSA}{Sensor, Observation, Sample and Actuator}
\newacronym{SLR}{SLR}{Systematic Literature Review}
\newacronym{BIM}{BIM}{Building Information Modeling}
\newacronym{TLO}{TLO}{Top-level Ontology}
\newacronym{PLM}{PLM}{Product Lifecycle Management}
\newacronym{CT}{CT}{Cognitive Twin}
\newacronym{BFO}{BFO}{Basic Formal Ontology}
\newacronym{HCI}{HCI}{Human-computer Interaction}

\begin{document}

\begin{frontmatter}



\title{Ontologies in Digital Twins: A Systematic Literature Review}


\author[inst1]{Erkan Karabulut\corref{cor1}}
\ead{e.karabulut@uva.nl}
\cortext[cor1]{Corresponding author.}

\author[inst2]{Salvatore F. Pileggi}

\author[inst1]{Paul Groth}

\author[inst1]{Victoria Degeler}

\affiliation[inst1]{organization={University of Amsterdam},
            addressline={Science Park 904}, 
            city={Amsterdam},
            postcode={1098 XH}, 
            state={North Holland},
            country={The Netherlands}}

\affiliation[inst2]{organization={University of Technology Sydney},
            addressline={15 Broadway}, 
            city={Ultimo},
            postcode={2007}, 
            state={New South Wales},
            country={Australia}}

\begin{abstract}
Digital Twins (DT) facilitate monitoring and reasoning processes in cyber-physical systems. They have progressively gained popularity over the past years because of intense research activity and industrial advancements. Cognitive Twins is a novel concept, recently coined to refer to the involvement of Semantic Web technology in DTs. Recent studies address the relevance of ontologies and knowledge graphs in the context of DTs, in terms of knowledge representation, interoperability and automatic reasoning. However, there is no comprehensive analysis of how semantic technologies, and specifically ontologies, are utilized within DTs. This \gls{SLR} is based on the analysis of 82 research articles, that either propose or benefit from ontologies with respect to DT. The paper uses different analysis perspectives, including a structural analysis based on a reference DT architecture, and an application-specific analysis to specifically address the different domains, such as Manufacturing and Infrastructure. The review also identifies open issues and possible research directions on the usage of ontologies and knowledge graphs in DTs. 
\end{abstract}



 \end{frontmatter}


\section{Introduction}\label{sec:intro}

Cyber-physical systems of the last decade have transitioned from using traditional system models to using Digital Twins (DT)~\cite{tello2021digital}. One of the most relevant and distinguishing features of Digital Twins is the real-time connection between the physical and the virtual system. It enables a more sophisticated digital model, which recreates and can update a physical environment faithfully and on the fly, rather than at later stages after simulation or analysis.

Digital Twins are applied to various vertical industries, the most common being manufacturing, agriculture, and construction ~\cite{correia2023data}. Given the variety of application areas, there is no single commonly-accepted definition of a Digital Twin. Additionally, the concept is constantly evolving to reflect advances in the field \cite{d2022cognitive}. 
In line with the broad characteristics of Digital Twins, there is also a variety of approaches to design and develop such systems as and there is, currently, no consensus on specific engineering processes and related architectures for them. Nevertheless, certain architectural patterns, which are discussed later on in this paper, have begun to emerge. 

Moreover, considerable gaps can be found in the current structured understanding of data and flow representation and reasoning layers of Digital Twins. 
One of the most well-known and promising standards for knowledge representation and reasoning is semantic technologies and, in particular, ontologies. An ontology provides a formal machine-processable conceptualization of a given domain~\cite{guarino1995formal}, including entities, their types and relationships, normally implemented in standard languages. These languages take advantage of the Web infrastructure to enable interoperability at a global level and to support automatic reasoning.
As shown further in this paper, there is a growing popularity of employing ontologies in the DT systems among researchers and engineers. Ontologically-enriched Digital Twins are often called Cognitive Twins~\cite{firstcognitivetwin}. 

Given this popularity, a number of questions arise: how exactly are DTs employing ontologies being used? In which parts of the DT architecture are ontologies the most beneficial? How can one ensure the biggest gains from using these systems? What are the common barriers and limitations to overcome in utilizing ontologies in DTs? 

At present, there are no best practices that have been established that answer the above questions. Additionally, a deeper analysis that provides an overview of the current application trends and of the relationships with the different architectural patterns is missing.

In order to address these gaps, we conducted a Systematic Literature Review (SLR) of recent research in digital twins utilizing ontologies. We screened 460 papers and extracted 82 directly relevant papers. These papers have been discussed against a reference architecture that reflects the most common patterns in Digital Twins. Our analysis aims to exhaustively address ontologies in the different architectural layers, and includes an inter-layer analysis to provide a more comprehensive framework. A significant number of reviewed articles include an implementation of a knowledge graph\cite{hogan2021knowledge} in DTs using ontologies, hence, an additional brief analysis of such knowledge graphs implementations was carried out. We also considered an application perspective, looking at the DT domains in which ontologies are most commonly used. 

More holistically, we identified a number of discussion points and possible future research directions based on the review.

\textit{Structure of the paper.} Section \ref{sec:background-concepts} provides an overview of the background concepts, while Section \ref{sec:related} presents the related work, focusing on SLRs on DTs. Section \ref{sec:methodology} addresses methodological aspects.
The core part of the paper is composed of 3 sections that deal respectively with the reference architecture (Section \ref{sec:ref-arch}), the performed analysis (Section \ref{sec:analysis-results}) and  its discussion (Section \ref{sec:discussion}). Finally, Section \ref{sec:conclusions} concludes the paper by summarising the major findings.

\section{Background Concepts} \label{sec:background-concepts}

This review aims to identify and discuss the body of knowledge associated with the application of ontologies in Digital Twins. This section addresses these two main concepts in order to provide a self-contained concise overview of the relevant background for understanding the subsequent review. 

\subsection{Digital Twins} \label{sec:background-dt}

\gls{DT} is a term that has become popular especially over the past 5 years. The term is used in multiple disciplines and contexts other than Computer Science, including, among others, several different sub-disciplines in engineering, business and healthcare (see Section \ref{sec:domain}). Different definitions have been used over the past 2 decades. D'Amico et al. \cite{d2022cognitive} in their \gls{SLR} have identified 11 different definitions. The first definition, without explicitly mentioning the term DT is given by Michael Grieves in 2002 \cite{grieves2002}, as the conceptual ideal for \gls{PLM}: \textit{``PLM is an integrated, information driven approach to all aspects of a product’s life from its design inception, through its manufacture, deployment and maintenance, and culminating in its removal from service and final disposal."}. In the presentation, separation of virtual and real space and the bi-directional communication in between is emphasized as the one of the main characteristic of a DT. 

DTs can be used to achieve different goals, such as physical space monitoring, optimizing decisions made by a physical system/asset, and predictive maintenance. Although the definition of DT slightly evolved over the time, the bi-directional communication in between the physical and virtual space remained as one of the distinctive features of a DT. A more recent definition given by Grieves and Vickers in 2017 in \cite{grieves2017digital} is: \textit{``Digital Twin is a set of virtual information constructs that fully describes a potential or actual physical manufactured product from the micro atomic level to the macro geometrical level. At its optimum, any information that could be obtained from inspecting a physical manufactured product can be obtained from its Digital Twin"}. Besides production in manufacturing, DTs are now also created for countries \cite {akroyd2021universal}, plants \cite{skobelev2020development}, and construction \cite{khan2022platology} just to name a few. 

\begin{figure}[tb]
    \centering
    \includegraphics[width=0.45\textwidth]{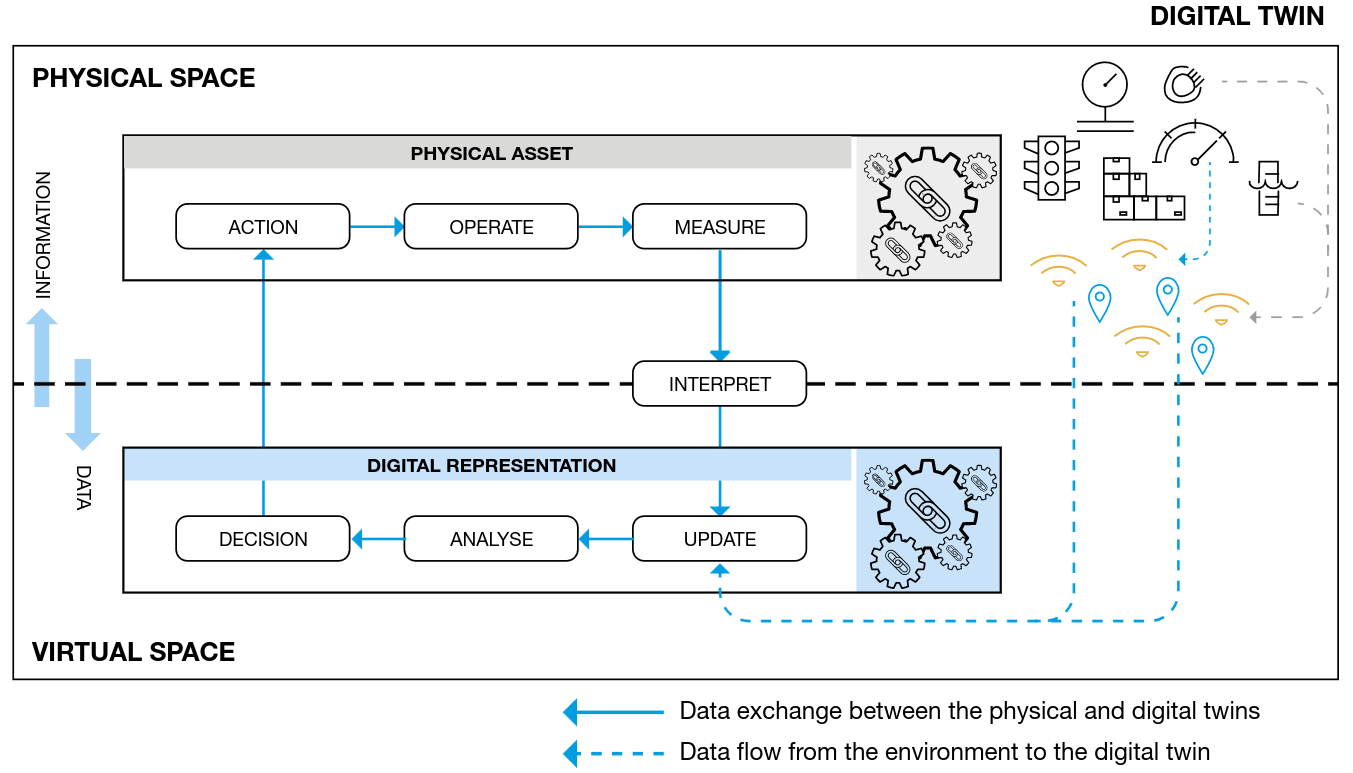}
    \caption{A figure from \cite{tello2021digital} illustrating the DT technology more comprehensively.}
    \label{fig:tello-dt}
\end{figure}

Figure \ref{fig:tello-dt}, taken from \cite{tello2021digital}, shows a more comprehensive, domain-agnostic definition for DTs. The main difference in such a view is the interpretation part, where data from the physical space is explicitly converted into a format that is processable as part of the virtual space. 

In spite of the fact that all of the DT definitions given in \cite{d2022cognitive} contain a physical asset, DTs are also created for abstracted entities. As an example, Parmar et al. \cite{parmar2020building} discuss building a DT for an organization, which also includes abstracted concepts related to companies. As the use of DTs becomes more widespread, the definition of DT is likely to keep evolving accordingly. 

Another concept frequently associated with DT is \gls{DS}. DS simply refers to a DT without any communication from the digital environment back to the physical environment. Finally a third term which is especially relevant in the context of this review is \gls{CDT} \cite{firstcognitivetwin}. It was defined by Ahmed El Adl in 2016: CDT \textit{"is a digital representation, augmentation and intelligent companion of its physical Twin as a whole including its subsystems and across all of its life cycles and evolution phases"}. CDTs are also referred as \gls{CT} \cite{lu2020cognitive} or Cyber Twins \cite{bamunuarachchi2020cyber}. Later definitions of CDTs (e.g., \cite{lu2020cognitive}) include explicitly Semantic Web technology, such as ontologies and knowledge graphs, as part of the CT technology. Although a Cyber Twin is not synonymous with CT, it still incorporates semantic technologies in a DT, while also considering Industry 4.0\cite{lasi2014industry} specific data management issues.    

\subsection{Knowledge Representation and Ontology} 

The term ``ontology" has its origin in philosophy, where, in a context of metaphysics, is used to deal with the ``nature of being". Although maintaining its original philosophical meaning, ontology has progressively evolved towards a more generic interpretation, as it is largely used in fact to express a formal conceptualization of a given domain.

More recently, ontology has been enabled also in computer science to create and work with formal machine-processable specifications of a given domain \cite{guarino1995formal}, often referred to as ``semantics". Ontologies became a key notion in the field of Knowledge Engineering \cite{brewster2004knowledge} . Its popularity in Computer Science consistently increased with the growth of the Semantic Web \cite{berners2001semantic}, which adopts the Web infrastructure to establish global identifiers. Indeed, unique identifiers allow a more sophisticated approach to interoperability, that can be established at a semantic level (Semantic Interoperability \cite{obrst2003ontologies}), as well as to data management and re-use within rich knowledge spaces \cite{li2013ontology}.  

The effective application of ontology within modern systems has been further fostered by the availability of specialised languages \cite{pulido2006ontology} (e.g. RDF  and OWL), most of which have been standardised by W3C\footnote{https://www.w3.org}. Such languages provide capability in terms of inference and automatic reasoning \cite{sirin2007pellet}\cite{shearer2008hermit}, and allow the establishment of semantically enriched data ecosystems, such as Linked Data \cite{bizer2011linked} and Open Data \cite{jain2010ontology}.

Ontologies normally work in the background of final systems and their value becomes even more relevant in distributed environments, where they typically contribute in the support of machine-to-machine interaction. However, ontologies may be considered a valuable asset also to support functionalities and representations in a generic context of \gls{HCI} \cite{costa2021ontologies}.  

The popularity of ontologies has progressively increased in the past two decades. The intense research activity within the community has resulted in a relatively consolidated technology that is being applied in a broad range of disciplines and application domains to solve real world problems. Typical examples of a successful application are in biology \cite{hoehndorf2015role}, medicine \cite{manika2018application}, system engineering \cite{dermeval2016applications} and manufacturing \cite{jia2023simple}.

\section{Related Work}\label{sec:related}

This section aims to provide a concise overview of related reviews that also focus on DTs and ontologies. 

A search on Scopus\footnote{https://www.scopus.com/home.uri} in titles, abstracts and keywords with a composed query resulting from the combination of "digital twin", "ontology" and "review" returned only 5 results. 20 more SLRs identified as a result of applying the methodology described in section \ref{sec:methodology}. Only 3 out of 25 of them have any analysis of ontologies when used in DTs. Those contributions are summarized in this section together.

D’Amico et al. \cite{d2022cognitive} performed an \gls{SLR} that includes 59 articles on \gls{CDT} (CT as per authors' statement) in the maintenance context. The analysis of Digital Twins assumes 5 different categories: purpose, communication, knowledge representation, computation and microservices. Knowledge Representation is relevant and directly connected to this work. Authors report that 28 of the selected articles adopt ontologies explicitly, with 5 of them referred to be \gls{TLO}. In order to improve interoperability, the reviewed articles benefit from standardized architectures, ontologies such as \gls{SSN}\cite{ssn-sosa}, or international standards, such as ISO\footnote{https://www.iso.org/home.html}. Another relevant aspect that is analyzed under the knowledge representation section is the type of database employed. The authors state that 3 types of databases have been commonly used (relational, non-relational and graph databases). RDF\footnote{https://www.w3.org/RDF/} and OWL\footnote{https://www.w3.org/OWL/} have been found as the most common data formats to store data as a knowledge graph.

Correia et al. \cite{correia2023data} carried out an SLR focusing on data management aspect in DTs. Results related to interoperability and data integration are especially relevant in the context of this work. The authors analyzed interoperability in DTs under 3 categories: data interoperability, semantic interoperability and interoperability in the communication level. On the semantic level, domain ontologies are used to provide semantic interoperability as well as for the communication in between different DTs in the same domain. As one of the data integration solutions, the authors found that modeling domain knowledge with an ontological layer in the architecture is also a common approach. Another part of the analysis was to understand for which domains the DT solutions were proposed in the reviewed articles. Industry 4.0, Smart Cities and Healthcare domains are found to be the most common application areas of DTs. In terms of application domains, the mentioned review presents results in line with our findings (Section \ref{sec:domain}). 

Shishehgarkhaneh et al. \cite{baghalzadeh2022internet} conducted an SLR specific to construction. The goal of the SLR is to understand how \gls{BIM}, DT and \gls{IoT} technologies are adopted in the construction industry. Although ontologies are not an explicit topic of review, authors identified the concept of ``ontology" as one of the most prominent in the reviewed articles. Authors state that ontologies have not yet been developed to address diverse and multi-context construction workflows. Our review has pointed out multiple ontologies being used for different aspects in the construction industry. However, we were also unable to identify concrete application of ontologies to address construction workflows. 

Although it is not an SLR, we have also found D’Amico et al.'s earlier work \cite{d2021top} highly relevant as they are also using the same search query as in our review, "digital twin" and "ontology". Authors briefly review existing scientific papers that use ontologies in the scope of a digital twin and found out that by the time the paper is written, a limited number of articles mention using an ontology and only a few mentioned using a \gls{TLO}-based approach. Finally, a TLO-based DT conceptual model is proposed for maintenance operations.

To the best of our knowledge at the time of writing this review, there are no SLRs that exhaustively deal with the adoption of ontologies in DTs. Our SLR differs from the existing work by solely focusing on how existing DT solutions benefit from ontologies. Based on the reference DT architecture given in Section \ref{sec:ref-arch}, this review investigates in which layers of a DT architecture ontologies are used, and the role of ontologies for each layer is identified (See section \ref{sec:structural-analysis}). 

It has been found that a single ontology might include concepts that belong to the multiple layers of a DT architecture. Therefore an inter-layer analysis is carried out especially to discover how ontologies play the role of a semantic interface in between layers (Section \ref{sec:inter-layer-analysis}). Some of the articles reviewed also construct a knowledge graph that is mostly based on a domain ontology. Although it was not among the initial analysis goals, a brief analysis of knowledge graph implementations in DTs is also performed (See sections \ref{sec:results-knowledge-graphs} and \ref{disc:knowledge-graphs}). Lastly, a domain-level analysis is performed to understand in which domains the ontologies are used most commonly in the scope of DTs.

\section{Methodology and Approach} \label{sec:methodology}

In order to better understand how ontologies are used in the scope of DTs, we performed a systematic literature review. This section explains the methodology of the review including how the articles are selected and analysed. The guideline published by Kitchenham et al. for performing a systematic literature review in Software Engineering \cite{kitchenham2007guidelines} has been taken as a reference for the review process. Both intermediary and end results of applying the methodology described in this section is made publicly available \cite{karabulut_erkan_2023_8172341}.

\subsection{Identification and Initial Selection of Research}

\begin{figure}[htp]
    \centering
    \includegraphics[width=0.47\textwidth]{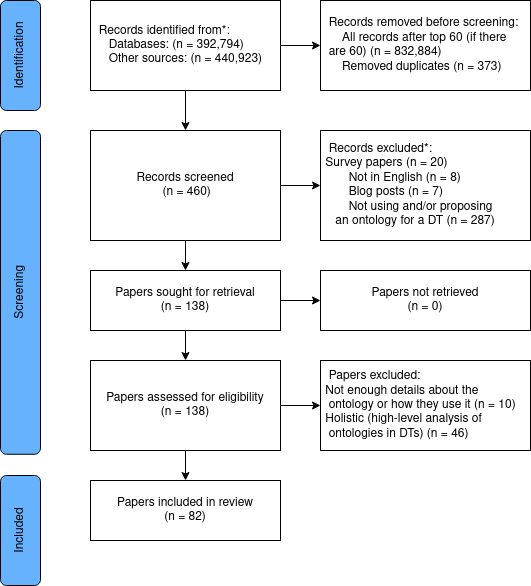}
    \caption{PRISMA Workflow showing the identification and screening processes for the SLR.}
    \label{fig:prisma-workflow}
\end{figure}

\textit{Data sources and search query.} 5 different relevant research databases (ACM Digital Library, arXiv, IEEExplore, ScienceDirect, SpringerLink) and 3 search tools (Google Scholar, Semantic Scholar, Zeta Alpha - AI Research Navigator) have been considered. 

Relevant papers have been retrieved  by the following queries: i) "digital twin ontology", ii) "digital twin" AND "ontology". Data was collected in the period of March/April 2023.

\textit{Initial retrieval.} Figure \ref{fig:prisma-workflow} shows the PRISMA workflow \cite{Pagen160} which summarizes the study identification and the screening process. As the initial number of results is considerably high (832,884), only the first 60 studies as ranked by the considered portal are retrieved. This is in line with empirical observations that show the relevancy to be negligible after a certain threshold \cite{o2015using}. In this specific case, the threshold (60) has been decided by skimming. However, not all data sources returned more than 60 results and, finally, 833 papers overall were retrieved. After removing 373 duplicates, 460 papers have been selected as an outcome of the initial screening process.

\textit{Exclusion/inclusion criterion and relevancy check.} Only research articles that explicitly propose a use of an ontology in the scope of a DT are included in this survey. Relevancy of the papers are decided in two analysis rounds. The first round focuses on the focus of the paper by considering title, abstract and keywords; additionally, skimming was performed to further assess the consistency in scope. Only articles written in English were considered. 8 papers only had an abstract in English. 7 of the results were blog posts and 20 of them were surveys which were considered not relevant as this SLR focuses on research articles only. 287 of the papers were either discussing Digital Twins or Ontologies, but not their nexus, or were not actually dealing with ontologies in the scope of a DT. Finally, 138 papers were assessed in detail in the second round. 10 of them did not provide enough detail about the ontology used or the benefit provided by ontology adopted. 46 of the papers were discussing the use and benefits of ontologies in DTs with a holistic short view, rather than actually utilizing ontologies or proposing a way to utilize ontologies in DTs. This resulted in 82 papers in total included in this survey.

\subsection{Overview of the selected research and analysis} \label{sec:methodology-analysis}

\begin{figure}[htp]
    \centering
    \includegraphics[width=0.45\textwidth]{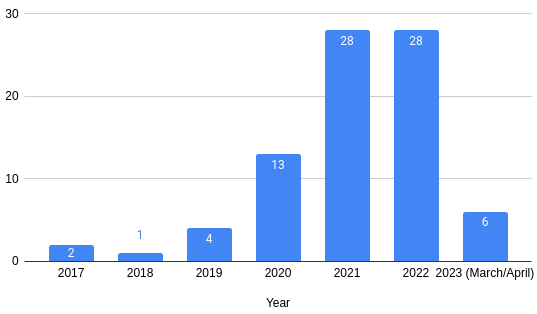}
    \caption{Number of relevant publications per year.}
    \label{fig:publications-per-year}
\end{figure}

Figure \ref{fig:publications-per-year} shows relevant publications per year. The number of publications increased almost 3 times over the period. This trend is not limited to ontologies only, but also applies to semantic technologies in general and knowledge graphs (see Section \ref{sec:analysis-results}).

Our analysis has been structured according to a reference architecture composed of different logical layers. Section \ref{sec:ref-arch} describes such a reference architecture in detail. 

In the analysis process, we have identified conceptual and functional patterns of ontologies to be matched with the logical layers of the architecture. As an example, if a concept in an ontology is used to describe a physical entity and that is specific to a domain, for instance floors of a building in building management, then the concept "floor" leads to a physical layer in the reference architecture in a specific domain (building management). As explained later on in the paper, a single ontology is often addressing concepts in the scope of more than  one layer. Indeed, ontology often acts as an interface in between layers. 

Those considerations led to the inter-layer analysis (section \ref{sec:inter-layer-analysis}). Additionally, the domain of each contribution is identified and a domain-based analysis is carried out accordingly. Lastly, even though it was not explicitly among the objectives, we have discussed also Knowledge Graphs because of their relevance and popularity in the reviewed articles. Tamašauskaitė et al. \cite{tamavsauskaite2023defining} has showed that utilizing ontologies is one of the common steps while constructing a knowledge graph. As it will be described in section \ref{sec:results-knowledge-graphs}, our review also shows that knowledge graphs are utilized in many of the reviewed research articles together with ontologies.

\section{Reference Architecture} \label{sec:ref-arch}

As far as we know, there is not a commonly accepted reference architecture for DT since authors tend to propose their own view of a DT architecture as part of their work. However, given the increasing popularity of DT, common architectural patterns are progressively emerging, although we cannot yet see a proper convergence of the different architectures. A common architecture is often perceived as a need within the community \cite{tao2018digital}.

Most architectures are structured in layers and are normally designed to reflect a seamless coexistence of a physical and a virtual space. That is the case of the architecture proposed by Ashtari et al.~\cite{ashtari2019architecture} that assumes two main layers (physical and cyber layer), while most architectures are structured in a more detailed way.

For instance, the architecture proposed by Souza et al. \cite{souza2019digital} extends the previous concept by adding a gateway between the physical and the digital layer. Similarly, in the work by Fan et al. \cite{fan2021digital} the authors integrate cyber-physical components with a human layer to address human-cyber-physical systems.
The architecture by Minerva et al. \cite{minerva2020digital} is structured in 4 layers, including data, integration, service and business, while Steindl et al. \cite{steindl2020generic} assumes a service-oriented architecture based on physical and virtual entities to support a given business logic. A full service-oriented approach structured in 5 different layers (Physical, Communication, Digital, Cyber and Application) is proposed by Aheleroff et al. \cite{aheleroff2021digital}.

Schroeder et al. \cite{schroeder2016visualising} propose five relatively classic layers (device, user interface, Web service, query, and data) integrated with a specific layer for augmented reality. A 5-layer architecture - i.e. Smart-Connection, Data-to-Information, Cyber, Cognition and Configuration - is proposed by Lee et al. \cite{lee2015cyber}. The six-layer architecture described by Redelinghuys et al. \cite{redelinghuys2020six} includes a double layer for physical twins (devices and data), local data repositories, an IoT Gateway, Cloud-based Information repositories and, finally, a layer for Emulation and Simulation.

An explicit cloud-based approach is adopted by Alam and Saddik \cite{alam2017c2ps}, with a duality between physical and
cloud cyber things, and by Gehrmann and Gunnarsson \cite{gehrmann2019digital}, which puts emphasis on Security.

In most mentioned architectures, data is implicitly addressed at different layers, without a specific data view.

\begin{figure}[htp]
    \centering
    \includegraphics[width=0.46\textwidth]{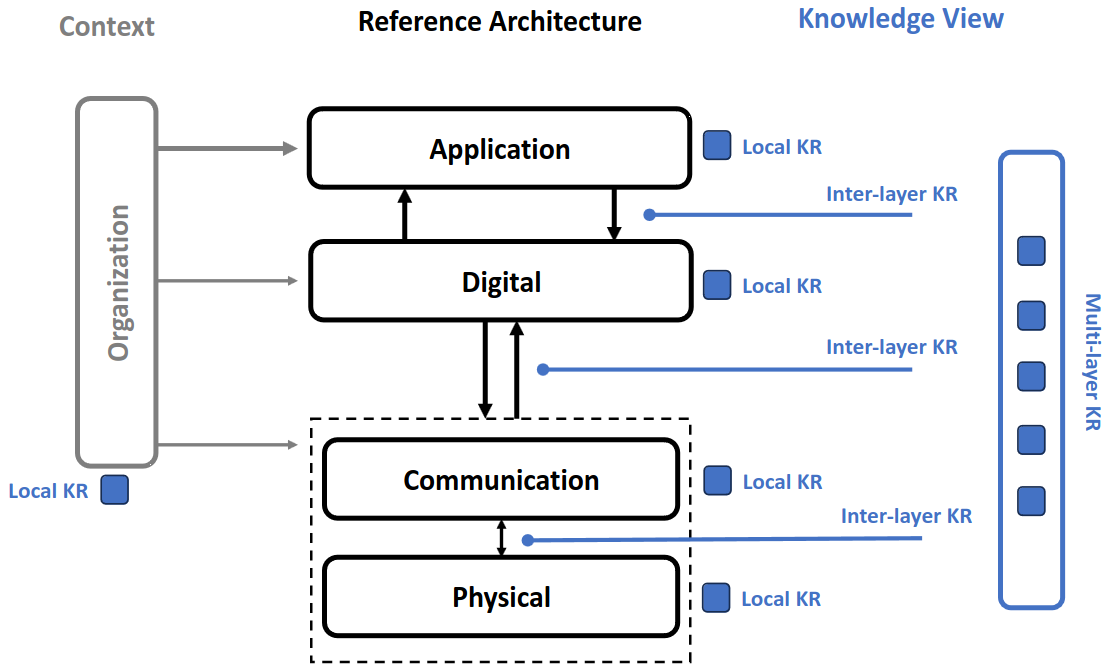}
    \caption{Reference Architecture.}
    \label{fig:architecture}
\end{figure}

The proposed literature review has been conducted looking at the reference architecture in Figure \ref{fig:architecture}:

\begin{itemize}
    \item \textit{Organizational Context}. Our analysis has been performed at a generic level without assuming any specific domain or context. We assume this virtual layer to reflect, represent or specify such specific aspects in a given context. While the main focus is on business and organizations, it may also include elements of system engineering at different levels.
    \item \textit{Logical Layers}. The reference architecture is structured in 4 different layers. The lower ones - i.e. \textit{Physical} and \textit{Communication} - aims to reflect the physical reality by addressing physical elements and their interaction respectively. This logical block is intuitively complemented by the \textit{Digital Layer}, which provides a digital representation of the physical world. Intuitively, the most abstracted layer (\textit{Application}) addresses application specific aspects and components. 
    \item \textit{Knowledge View}. We are assuming a fluid model for knowledge representation which assumes 3 different kinds of support: (i) \textit{local}, when the representation is in the specific boundaries of one single logical layer, (ii) \textit{inter-layer} to interface two contiguous layers and (iii) \textit{multi-layer} if involving two or more non-contiguous layers.
\end{itemize}

\section{Ontologies in Digital Twins} \label{sec:analysis-results}

This section presents the results of the analysis conducted. Table \ref{table:analysis-results} shows the list of reviewed papers with associated details, including ontology name, application domain, related architecture layer, and whether a given solution utilizes KGs or not. The section is structured in 5 different parts that deal respectively with (i) the value provided by ontologies in DTs, (ii) structural analysis by layer, (iii) inter-layer analysis, (iv) domain analysis and (v) Knowledge Graphs.

\subsection{Objectives}

There has been 4 different inter-related objectives observed in the reviewed articles: system/data modeling, semantic interoperability, (implicit) semantic relation extraction, and automated reasoning support. These objectives are either explicitly mentioned by the authors as the reason to employ ontologies, or in case it was not mentioned explicitly, we found the purpose of employing ontologies fitting to one or more of the mentioned objectives. Those objectives may be considered to be layer-agnostic, meaning they normally affect a system as a whole. 

\textit{System/Data modeling.} Based on our review, one of the common reasons for incorporating ontologies into DTs is to model the DT system and to integrate heterogeneous data from the various components of a DT~\cite{karabulut_erkan_2023_8172341}. Domain ontologies help modeling parts in a DT, and data structures to be stored or data packages to send other internal/external parts of a system. Since a domain ontology includes all the concepts that belong to a domain, if comprehensive enough, such an abstracted model can effectively drive developments. As an example, Zhang et al. \cite{zhang2017modeling} create a DT model for workshops utilizing a proposed domain ontology that consists of 3 main classes: \textit{ResourceInformation}, \textit{TaskInformation} and \textit{ProcessInformation}. A further development assumes the refinement of the main classes to include sub-classes and properties. 

\textit{Semantic interoperability.} It refers to understanding what a piece of data means when sent to a different sub-component in a DT. DTs can also co-exist and even cooperate e.g., to share learned parameters for a common task that is performed in multiple DTs \cite{lu2021cognitive}. Ontologies can provide this semantic understanding of the data across sub-systems or DTs. While domain ontologies are usually enough to establish semantic interoperability for the entities in the same domain and context \cite{liu2020research}, top-level ontologies can also play a role when used across domains~\cite{petrova2021digital}.

\textit{Semantic relation extraction.} Ontologies, especially when used to build knowledge graphs and supported with sensor data, can help extracting implicit semantic relations in DTs. Knowledge graphs are composed of instances of classes that are described in ontologies. Sensors are used to track the latest state of these instances, and rule extraction algorithms can be altered to work with knowledge graphs and sensor data to extract implicit semantic relations \cite{banerjee2017generating}. 

\textit{Reasoning facilitation.} A rather more generic reason to use ontologies is to facilitate automatic reasoning in the system. Most automated reasoners require data from diverse sources in a DT or across DTs, as well as semantic information about this data to be able to process it. Output of the reasoning is then propagated to respective components in accordance with the used ontology. Hoebert et al. \cite{hoebert2019cloud} use an ontology-based model of industrial robots and run reasoning algorithms to plan a set of actions to reach a certain goal of the system.  

\subsection{Structural analysis} \label{sec:structural-analysis}

Figure \ref{fig:concept-map} shows the number of publications that uses an ontology in the scope of a DT per layer of the reference DT architecture (inside the circles). The same figure also shows a pairwise analysis of ontologies that addresses more than one layer simultaneously. 63 out of 82 publications use ontologies to describe concepts that belong to the physical layer. 49 of the papers focus on the DT layer, 15 on the organization layer, 15 on the application layer and 5 on the communication layer. 

\begin{figure}[htp]
    \centering
    \includegraphics[width=0.48\textwidth]{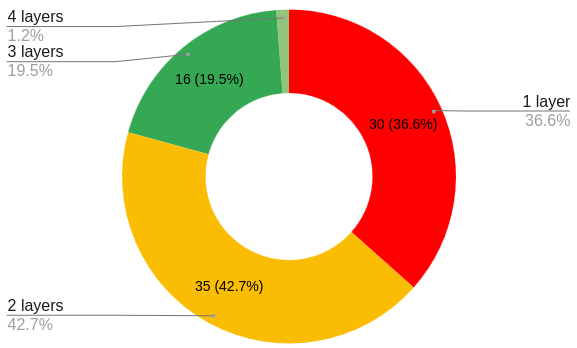}
    \caption{Number of articles utilizing ontologies in a DT that has concepts belongs to n (1-4) layers.}
    \label{fig:papers-by-ontology-layer-match}
\end{figure}

In most cases, ontologies have a multi-layer focus. Figure \ref{fig:papers-by-ontology-layer-match} shows the number of papers where ontologies include concepts from 1 or more layers. Ontologies in 30 out of 82 papers include concepts that belong to 1 layer only, where 23 of them are matched to physical layers. These ontologies can be considered as domain-specific, while the rest of the 52 ontologies found in the reviewed articles are more task- and/or application-oriented. 35 of the ontologies include concepts that belong to 2 layers, where majority of them are matched to physical-digital layers. 16 of the reviewed articles include ontologies where concepts belong to 3 layers and only 1 article found where the ontology include concepts from 4 layers. 

The following subsections explain how ontologies are specifically used in different layers of the reference DT architecture. A summary of the results are given in Table \ref{table:ontology-usage} where the column "Architecture Layer" refers to which layers do the concepts used in the mentioned ontology (or ontologies) corresponds to. 

\begin{table}[ht!]
\centering
\caption{Usage of ontologies in the different layers as per reference architecture.}
\rowcolors{2}{white}{whitesmoke}
\begin{tabular}{p{0.12\textwidth}p{0.30\textwidth}}
 \hline
        \rowcolor{floralwhite} Layer & Usage \\ \hline 
        Physical & physical entities, actions and processes \\
        Communication & protocols, access parameters \\
        Digital & generic DT concepts, real or abstract/derived digital terms, assets and operations \\
        Application & ranges between task-specific terms (e.g., CNC (Computer Numerical Control) cutting machine optimization app) to domain-independent application terms (e.g., top-level requirements validation app) \\
        Organizational context & production lines, facilities, client and order info, project management, bridging DT and non-DT parts \\
\end{tabular}
\label{table:ontology-usage}
\end{table}

\subsubsection{Physical layer}

Two different usages of ontologies are found that describe concepts in the physical layer. The first one is to utilize ontologies to describe physical components, their physical attributes, states in a system and the relation in between them. Skobelev et al. \cite{skobelev2020multi} use an ontology to describe physical parts of a plant, such as root, stem or leaf, and the ontology is then used to extract rules for decision making. Another example in manufacturing is to represent industrial machines or machine parts, personnel, or environmental conditions such as temperature and humidity using an ontology. Liu et al. \cite{liu2020research} developed a CNC machine tool ontology that includes concepts such as Material, Personnel, Device and Environment. The ontology is used to aggregate data from diverse sources.  

Secondly, ontologies are also used to represent physical actions or processes. Tuli et al. \cite{tuli2021knowledge} used CORA ontology \cite{balakirsky2017towards} to represent movements of an industrial robot. Nguyen et al. \cite{van2022toward} proposed an ontology model for tactile sensing devices that has concepts describing tactile events such as position, velocity and type of a touch event. 

\subsubsection{Communication layer}

On the communication layer, ontologies are used to represent communication protocols in between far-edge, edge, and more centralized units such as cloud stores, or different parts of a machining system, production line. Chevallier et al. \cite{chevallier2020reference} proposed a reference DT architecture for smart buildings and utilizes many ontologies including \gls{SOSA}~\cite{janowicz2019sosa} ontology. Authors utilized \textit{Procedure} subclass of SOSA to specify communication protocol used and its attributes such as IP address. Maryasin \cite{maryasin2019home} developed a home automation system ontology that contains communication network-level classes such as \textit{NetworkProtocol} class. 

\subsubsection{Digital layer (DT)}

Both generic DT ontologies and ontologies that are used to represent digital entities are included in this category. 3 different usages of ontologies have been identified on the digital layer. The first one relates with representing concepts that generically used in a DT. Duan et al. \cite{duan2020development} propose a domain-independent DT ontology that consists of 4 categories of concepts: entity-related, DT-entity related, DT system and application framework dimensions. None of the proposed concepts are domain-specific and can be used in any DT implementation. These ontologies can also be used as a top-level ontology for DTs. The authors use the ontology while creating a reference DT architecture. 

The second way of using ontologies is to represent digital assets and operations such as settings of a machine, input that goes to a machine or a software module, operating systems etc. Khan et al. \cite{khan2022platology} created a construction DT ontology named ConDT ontology. It includes \textit{Data Resources} as a part of a construction DT. According to the ConDT, a data resource has a data source, data format, input method, database and an owner. A third way of using ontologies on the DT layer is to represent abstract (usually domain-specific) concepts in terms of digital data. Amar et al. \cite{amar2021knowledge} created an ontology for fault management and validates in a power plant scenario. The ontology has a \textit{Component} class which can have a \textit{Fault}, in a power plant. \textit{Component} has \textit{Sensors} which generates \textit{sensor\_stream\_data}. \textit{DataRule}s defined on the \textit{sensor\_stream\_data} to detect \textit{RootCause}s of faults. In this case, a fault is an abstract term to specify a data rule violation in a sensor of a component due to a root cause.

\subsubsection{Application layer}

The ontologies used in the application layer ranges between representing concepts that are specific to a certain task in a certain domain to more generic domain-independent application terms of a DT. Zheng et al. \cite{zheng2022semantic} introduce the requirements ontology for aircraft assembly systems and benefit from it while designing assembly processes. A set of ontologies are created to be used in construction domain in the scope of COGITO\footnote{https://cogito-project.eu/} project. Katsigarakis et al. \cite{katsigarakis2022digital} developed the COGITO ontology with 4 new modules which are then used to create a knowledge graph for construction projects: facility, process, resource and quality modules. Poudel et al. \cite{poudel2022integrated} developed a more generic ontology for manufacturing to represent manufacturing resources (e.g., machines), capabilities of the resources, and manufacturing processes. The ontology is used to automatically match resource capabilities to manufacturing processes.

\subsubsection{Organizational context}

DTs are also created for either entire organizations or parts of an organization. In harmony with this, ontologies created for these DTs either include broader concepts to cover operations and assets in an organization, or concepts that relate with a certain entity which the organization creates a DT for. 

Three different usages of ontologies in a DT are identified in an organizational context. The first one applies to representing production lines, facilities, received orders or client information of an organization. Rožanec et al. \cite{rozanec2020towards} proposed the term \textit{Actionable Cognitive Twin (ACT)} which is very similar to Cognitive Twins introduced in Section \ref{sec:background-dt}, however with more concrete PT interaction definitions. In a later work \cite{rovzanec2103actionable}, the authors proposed a manufacturing ontology based on \gls{BFO}~\cite{arp2015building} to be used in ACT. The ontology focuses on manufacturing concepts that are related to production planning and demand forecasting such as \textit{manufacturing process}, \textit{stock order}, \textit{production line}, \textit{production plant} and \textit{organization}. This is a good example of using ontologies in an organizational context in manufacturing.

Another type of utilization of ontologies in an organizational context is to track ongoing, long-lasting projects of organization(s). Münker et al. \cite{munker2021online} proposed Internet of Construction On-Site Ontology (IoC-OSO) that re-uses concepts from 5 other ontologies, including Domain Ontology for Construction Knowledge (DOCK 1.0) \cite{el2013domain}. DOCK includes semantic concepts that can be used to refer projects, their stages, states and life-cycle. IoC-OSO is used for resource allocation to construction processes. 

Third, Ariansyah et al. \cite{ariansyah2022enhancing} focuses on the problem of connecting DTs with other software systems used in an organization such Computerized Maintenance Management Systems (CMMS) or Enterprise Resource Planning (ERP) systems. The authors propose an ontology to establish semantic interoperability between DT and non-DT software.  

\subsection{Inter-layer analysis} \label{sec:inter-layer-analysis}

\begin{figure}[htp]
    \centering
    \includegraphics[width=0.48\textwidth]{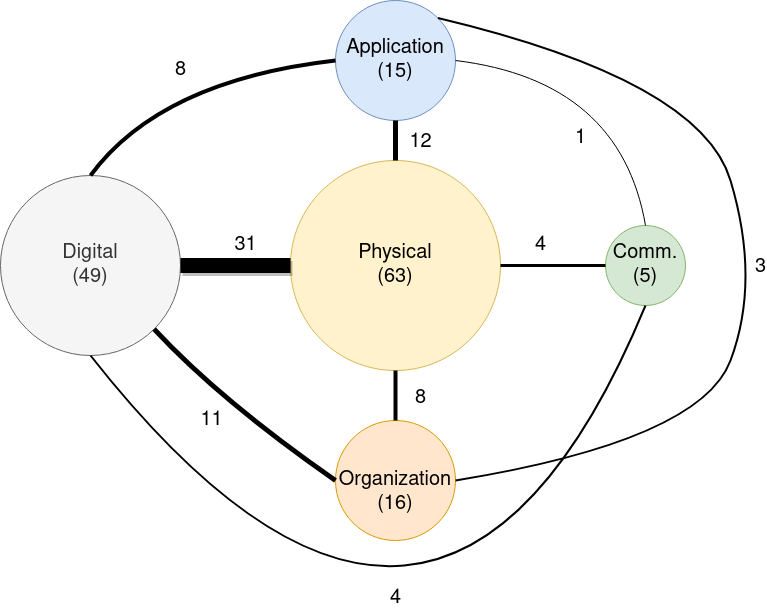}
    \caption{Pairwise analysis of ontologies used for concepts in different layers of a DT.}
    \label{fig:concept-map}
\end{figure}

As reported in Table \ref{table:analysis-results}, 30 of the publications use ontologies to describe concepts from 1 single layer only, while the majority of the papers, 52, use ontologies to describe concepts from multiple layers. Figure \ref{fig:concept-map} shows a pairwise analysis of ontologies used in the scope of DTs, that includes concepts from different layers. Some ontologies include concepts from 3 or more layers.

31 of the ontologies include concepts that belong to both physical and digital layers. This shows that physical and digital layers are semantically the most connected layers. On the other side, there is no ontology that describes concepts from the communication and the organization layer at the same time. Therefore these 2 layers are not connected at all. 

Physical layer is the one that has the most connections to other layers, while the communication layer has the least connections. Application and communication layers are most connected to the physical layer, while the organization is most connected to the digital layer.  

\subsection{Domains} \label{sec:domain}

\begin{figure}[htp]
    \centering
    \includegraphics[width=0.48\textwidth]{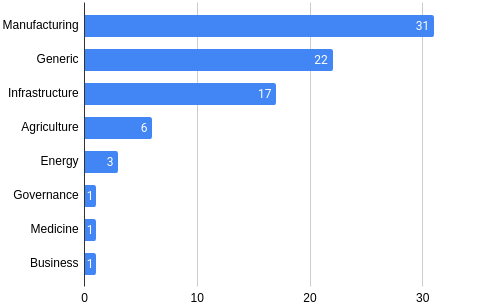}
    \caption{Number of papers by domain.}
    \label{fig:papers-by-domain}
\end{figure}

This section presents an analysis from an application domain perspective. Table \ref{table:analysis-results} includes domains for each of the paper in the \textit{Domain} column. An aggregated view of the papers by domain is given in Figure \ref{fig:papers-by-domain}.

The number given as \textit{Generic} is the sum of the papers with \textit{digital twin, IoT, smart home, materials science, IT, smart city, IT security} domains (see Table \ref{table:analysis-results}). \textit{Agriculture} refers to \textit{smart farming} and \textit{smart fisheries}. Lastly, \textit{Infrastructure} refers to \textit{building management, construction, public infrastructure} and \textit{cultural heritage} domains. 

D'Amico et al. \cite{d2022cognitive} in their SLR on cognitive DTs in maintenance context, have also reviewed papers based on application domain. Similar to their findings, ontologies in DTs are also mostly used in the Manufacturing domain, which is followed by Generic and Infrastructure domains. There has been only 1 paper found for \textit{Governance}, \textit{Medicine} and \textit{Business} domains. Another SLR on DTs performed by Correia et al. \cite{correia2023data} from a data management perspective. The SLR includes both domain and subdomain classification for the reviewed papers and the authors found that there are more papers in \textit{Smart Manufacturing} subdomain.

\subsection{Knowledge Graphs} \label{sec:results-knowledge-graphs}

As defined by Hogan et al.~\cite{hogan2021knowledge}, a knowledge graph is \textit{"a graph of data intended to accumulate and convey knowledge of the real world, whose nodes represent entities of interest and whose edges represent potentially different relations between these entities"}. Tamašauskaitė et al. \cite{tamavsauskaite2023defining} defined the steps to construct a knowledge graph, and included ontology construction as one of the steps. 17 of the reviewed articles build knowledge graphs using ontologies (see Table \ref{table:analysis-results} where the column \textit{KG} refers to whether a paper includes a knowledge graph implementation or not). Therefore, even though knowledge graphs are not the focus of this SLR (and the search query is not inclusive for knowledge graphs), this section is dedicated to a short analysis of how knowledge graphs are used in the reviewed articles. 

3 ways of utilizing knowledge graphs in the scope of a DT are identified. One common way of using knowledge graphs in the reviewed articles is to benefit from the graph structure and run queries on the node end edge properties using ontological terms to extract information, where each node includes metadata about DT components. Banerjee et al. \cite{banerjee2017generating} created knowledge graphs for industrial production lines and uses Path Ranking algorithm to extract semantic relations which are not as explicitly exist in the knowledge graph. 

The state of each component in a DT is frequently associated and stored together with the knowledge graph. Chukkapalli et al. \cite{chukkapalli2021cyber} creates a knowledge graph from fused sensor data (as opposed to metadata about the system components) in the DT of a smart farming use case. In this way, the latest state of the DT is always kept within the knowledge graph. Later, the knowledge graph is used to detect anomalies in the sensor data. 

Lastly, another way of utilizing knowledge graphs in DTs is for integrating heterogeneous data from multiple data sources. Proper et al. \cite{proper2021towards} developed an ontology-based DT for IT infrastructures of organizations. The authors mentioned that there is diverse data streams coming from IT Governance Processes, IT Management Processes and Organizational IT Assets. An ontology named Governed IT Management (GITM) ontology is described and a knowledge graph-based approach is built to handle unify the heterogeneous data.

\section{Discussion} \label{sec:discussion}

We now discuss the main outcomes of our review.

\subsection{Ontologies in Different Layers of a DT}

A DT by definition (Section \ref{sec:background-dt}) includes intelligent operations based on the application domain, which distinguishes it from a DS, and this requires a semantic understanding of the data. Ontologies can provide this semantic understanding. In most of the cases, ontologies used in DTs include concepts that belong to multiple layers (Sections \ref{sec:structural-analysis} and \ref{sec:inter-layer-analysis}) based on our reference architecture (Section \ref{sec:ref-arch}). In this way, ontologies also act as a semantic interface in between different layers. 

There are more articles that use an ontology to represent concepts in the physical layer than others (see Figure \ref{fig:concept-map}). One possible reason could be that the physical layer is more tied to respective domain. As an example, an application that checks if certain requirements are satisfied could be required in any domain listed in Figure \ref{fig:papers-by-domain}. Therefore many of the terms in this application would be similar across domains, hence less work is needed to formalize the concepts. However, components of the physical layer tend to be more domain-specific. This could also explain the high number of ontologies (31) that describe concepts from both physical and digital layer.

DTs can be created for single entities (an industrial machine), a set of entities in the same context (machines in a production line), or even entities that are completely in different domains (e.g., Akroyd et al. \cite{akroyd2021universal} creates a universal DT for UK). In the last two cases, DTs will have heterogeneous data from multiple sources. To solve this issue using ontologies, a single domain-specific ontology would suffice to unify the data for the second case. In the third case, the data might be representing concepts that belong to different layers in a DT architecture. Matching each domain ontology to a top-level ontology~\cite{d2021top} is among one of the popular solutions that can be used in the third case. 

\subsection{Application Domain}

Similar to results of other recent SLRs on DTs \cite{d2022cognitive,correia2023data}, there are more papers published in the manufacturing domain that use an ontology in the scope of a DT than in the other domains (see Figure \ref{fig:papers-by-domain}). A simple query of \textit{``ontology" AND ``digital twin"} on scopus gives 141 results in \textit{Engineering} (the one that is most related with manufacturing and infrastructure among the subject areas on Scopus search results), 16 in \textit{Energy}, 15 in \textit{Business, Management and Accounting} and 3 in \textit{Medicine}. When compared with the mentioned SLRs on different aspects of DTs, usages of ontologies in DTs across domains are proportionate to the number of articles published on DTs in general. 

\subsection{Knowledge Graphs}\label{disc:knowledge-graphs}

17 out of 82 papers utilized knowledge graphs. DTs, by nature, are closed systems with limited number of components where each component is somehow in an interaction with other, mostly neighbouring, components. Knowledge graphs can reflect these interactions semantically. As presented in Section~\ref{sec:results-knowledge-graphs}, knowledge graphs are actively used as a data store to store both metadata about the DT parts, as well as current state of each part based on sensor data. Knowledge graphs are then queried to extract system parts with certain patterns or simply to get latest system or component state. Therefore, knowledge graphs are also frequently used together with other decision support and reasoning algorithms. However, knowledge graphs so far used only as a metadata or state store, rather than as part of a reasoning process, e.g., guiding a reasoning algorithm based on the extract knowledge. We expect knowledge graphs to be more involved in future DT implementations, not only as a data storage but also actively as a part of reasoning algorithms.

\subsection{Multiple ontologies used in a single layer} \label{disc:mult-ont-in-single}

In Section \ref{sec:inter-layer-analysis}, we showed that some ontologies include concepts that belong to multiple layers in the DT architecture. However, in some cases multiple ontologies are also used in a single layer, especially in the case of relatively bigger DT systems. An example would be the smart city use case where multiple ontologies used together to semantically represent an entire city or even a country~\cite{akroyd2021universal}. In these cases, ontologies are integrated either by matching some of the common terms (or creating common terms) and using namespacing, or utilizing a comprehensive enough top-level ontology \cite{jiang2023intelligent, petrova2021digital, akroyd_harper_soutar_farazi_bhave_mosbach_kraft_2022}. 

Besides the semantic aspect of integrating ontologies, there is also the technical aspect of how and where to store and retrieve ontologies or knowledge graphs built using ontologies as well. Apache Jena\footnote{https://jena.apache.org/} is one of the tools that is used in linked data applications, and also in some of the papers reviewed~\cite{zhao2021cutting, zhang2017modeling}, to parse ontologies and also to store them in RDF stores such as TDB\footnote{https://jena.apache.org/documentation/tdb/index.html}, an RDF storage. Besides triple stores, property graphs such as Neo4j\footnote{https://neo4j.com/} is also among popular choices to store ontologies and knowledge graphs~\cite{hoebert2019cloud}. 

\subsection{Distributed Digital Twins}

A research topic that is just started to be studied by researchers is to have multiple DTs co-exists in a same context. Poudel et al. \cite{poudel2022integrated} created a framework with a pool of DTs for various manufacturing devices, where a decision maker unit tries to optimize configurations of DTs. Although we did not encounter it while performing this SLR, federated learning approach also seems promising when having distributed digital twins. An example in manufacturing would be to have multiple of the same or similar machines that perform a similar task and optimize its own configuration while running. Each machine then shares the learned parameters with DTs of other machines. In this way, there would be no need to have a central decision maker unit, and instead each DT has its own decision maker which can evaluate the learned parameters from other DTs. Semantic technologies such as ontologies and knowledge graphs are also not yet studied in the case of having distributed digital twins. A possible research direction would be to investigate technical and semantic integration of ontologies and knowledge graphs in the case of having multiple DTs co-exist. What are the pros and cons of having a single or many knowledge graphs for a same type of PT or different types of PTs? 

\subsection{Knowledge engineering and ontology re-use}

64 out of 82 of the reviewed articles proposed a new ontology. Only 19 of the 64 articles either re-used some concepts from existing ontologies or matched the newly proposed ontology to top-level ontologies. More than half of the articles did not mention creating a source file for the ontology and sharing it openly. This shows that the common problem of re-using ontologies in semantic web also exists for ontologies in DTs as well. One reason that we think it could be DT specific is that many ontologies are created for or based on a specific task or specific aspect to better or optimize (e.g., an ontology for energy usage of a particular industrial machine) and therefore can not be generalized to a domain. Matching newly proposed ontologies to top-level ontologies, sharing a source code openly on open access ontology databases can help alleviate the issue.  

\section{Conclusions}\label{sec:conclusions}

Digital Twins are becoming increasingly popular across many domains as research shows clear benefits in monitoring, decision support and reasoning tasks besides others. Semantic technologies are also being incorporated into DTs for better knowledge representation and to facilitate reasoning. Such digital twins are often called Cognitive Twins (CT). This SLR includes an analysis of 82 scientific papers that use an ontology in the scope of a DT. Its key findings are:

\begin{itemize}
    \item Ontologies are mostly used to represent concepts in the physical layer, which is interpreted as the physical layer being more tied to the respective domain. 
    \item 30 out of 82 reviewed articles have ``domain-specific'' ontologies, which at describe concepts from 1 layer, while 52 articles have ``application/task oriented'' ontologies where concepts stem from multiple layers. 
    \item Both DT and CT implementations and advancements are led by and often limited to the Manufacturing and Infrastructure domains.
    \item Ontology re-usability issues in semantic web persists for DTs as more than half of the reviewed articles did not re-use an existing ontology, or matched their proposed ontology to a top-level ontology.
    \item Knowledge graphs are becoming increasingly popular in DTs, due to their expressiveness of semantic relations and fast query capabilities. 
\end{itemize}

It has been only a couple of years since semantic technologies have been used in the scope of DTs. We believe that the capabilities offered by ontologies and knowledge graphs have yet to be fully leveraged by DTs. Below are some of the promising future research directions based on this systematic literature review: 

\begin{itemize}
    \item \textit{Integration of ontologies into DTs.} This SLR does not cover in detail the manner in which ontologies are integrated into DT knowledge bases both semantically and technically. Analysing DT specific requirements of the integration process will help researchers and practitioners to employ ontologies faster. 
    \item \textit{Widespread adoption of ontologies in DTs across domains.} This SLR showed that cognitive twins are so far adopted mainly in Manufacturing and Infrastructure domains. However, we believe that there cognitive twins can bring enormous value to other domains where twinning technology is applied. 
    \item \textit{Knowledge graph as a state graph.} Knowledge graphs are mostly used for storing metadata about DT components. However they can also be used as a state graph when combined with aggregated sensor data. This can help reducing further data processing time and can facilitate reasoning process.
    \item \textit{Knowledge graphs as part of reasoning process.} Besides being used as a data store, we believe that knowledge graphs in DTs can also bring great value to reasoning processes. They have the potential to guide reasoning algorithms, e.g., to decide where in the system should the reasoning be performed. 
\end{itemize}

We hope that this SLR can help researchers and practitioners to understand how ontologies are currently being used in digital twins and what are some of the future research directions. 

\section*{Acknowledgments}
This work has received support from The Dutch Research Council (NWO), in the scope of Digital Twin for Evolutionary Changes in water networks (DiTEC) project, file number 19454.

\onecolumn
\begin{longtblr}[  caption = {List of reviewed papers with proposed/used ontology name, domain, corresponding architecture layer and KG utilization.},
  label = {table:analysis-results},
]{
  colspec = {p{0.07\textwidth}p{0.04\textwidth}p{0.21\textwidth}p{0.21\textwidth} p{0.10\textwidth}p{0.15\textwidth}p{0.05\textwidth}},
  rowhead = 1,
  hlines,
  cells = {font = \fontsize{9pt}{11pt}\selectfont},
  row{even} = {whitesmoke},
  row{odd} = {white},
  row{1} = {floralwhite},
}
        Reference & Year & Proposed ontology & Re-used ontology & Domain & Architecture Layer & KG \\ \hline
        \cite{zhang2017modeling} & 2017 & Ontology of workshop manufacturing system & - & Manufacturing & DT & No \\ 
        \cite{banerjee2017generating} & 2017 & Ontology for manufacturing & - & Manufacturing & Physical, DT, Organizational & Yes \\ 
        \cite{steinmetz2018internet} & 2018 & Extension to IoT-Lite ontology & - & IoT & Physical, DT & No \\ 
        \cite{hoebert2019cloud} & 2019 & - & Rosetta\cite{patel2012enabling}, OntoBREP CAD \cite{perzylo2015ontology} ontologies & Manufacturing & Physical & No \\ 
        \cite{perzylo2019opc} & 2019 & OPC UA Ontology & OntoBREP CAD Ontology\cite{perzylo2015ontology} & Manufacturing & Physical & No \\ 
        \cite{david2019attaining} & 2019 & Manufacturing, Learning and Pedagogy Ontology & - & Manufacturing / Pedagogy & Physical, Application & No \\ 
        \cite{maryasin2019home} & 2019 & Home automation ontology & - & Smart Home & Physical, Communication, DT & No \\ 
        \cite{bao2022ontology} & 2020 & Machine parts ontology for manufacturing & - & Manufacturing & Physical & No \\ 
        \cite{barth2020systematization} & 2020 & Digital twin ontology & - & Digital Twin & DT, Organizational & No \\ 
        \cite{morgado2020mechanical} & 2020 & Mechanical testing ontology & - & Materials Science & Physical, Application & No \\ 
        \cite{liu2020research} & 2020 & CNC Machine ontology & - & Manufacturing & Physical & No \\ 
        \cite{petrova2021digital} & 2020 & Upper level city ontology for DT modeling & - & Smart City & Physical, DT & No \\
        \cite{bamunuarachchi2020cyber} & 2020 & A DT ontology & SSN and SOSA\cite{ssn-sosa} & Digital Twin & Physical, DT, Application & No \\ 
        \cite{skobelev2020development} & 2020 & Plant DT ontology & - & Smart Farming & Physical, DT, Application & No \\ 
        \cite{chevallier2020reference} & 2020 & - & ifcOWL (OWL for Industry Foundation Classes)\cite{pauwels2016express}, SSN, SOSA, BOT (Building Ontology Topology)\cite{botontology} & Building Management & Physical, Communication, Application & No \\ 
        \cite{skobelev2020multi} & 2020 & Ontology of plant DT & -  & Smart Farming & Physical & No \\ 
        \cite{bao2021product} & 2020 & Manufacturing product ontology & - & Manufacturing & Physical & No \\ 
        \cite{liu2020web} & 2020 & Manufacturing ontology & Re-uses concepts from SSN\cite{ssn-sosa} & Manufacturing & Physical, Application & No \\ 
        \cite{duan2020development} & 2020 & DT ontology & - & Digital Twin & DT & No \\ 
        \cite{singh2021data} & 2021 & An ontology for DT data management & - & Digital Twin & DT & No \\ 
        \cite{dai2021ontology} & 2021 & Geometric information ontology & STEP-NC machine tool control language\cite{step-nc} & Manufacturing & Physical, Communication & No \\ 
        \cite{bao2021ontology} & 2021 & Assembly workshop ontology & - & Manufacturing & Physical, Organizational & No \\ 
        \cite{goppert2021pipeline} & 2021 & - & MASON\cite{masonontology}, Brick\cite{brickontology} and BOT\cite{botontology} & Manufacturing & Physical, DT & No \\ 
        \cite{proper2021towards} & 2021 & Governed IT Management (GITM) Ontology & - & Governance / Management & Organizational & Yes \\ 
        \cite{ayinla2021semantic}  & 2021 & Offsite Manufacturing Production Workflow Ontology & - & Manufacturing & Physical, Application, Organizational & No \\ 
        \cite{zheng2021hierarchical} & 2021 & Mechanical products ontology & - & Manufacturing & Physical & No \\ 
        \cite{skobelev2021development} & 2021 & A Plant DT Ontology & - & Smart Farming & Physical, DT, Organizational & No \\ 
        \cite{tuli2021knowledge} & 2021 & - & CORA\cite{balakirsky2017towards}, SSN\cite{ssn-sosa} ontologies & Manufacturing & Physical & No \\ 
        \cite{akroyd2021universal} & 2021 & - & 10+ various domain ontologies & Digital Twin & Physical & Yes \\ 
        \cite{massel2021ontologies} & 2021 & A domain and a DT ontology for the energy domain & - & Energy & Physical, DT & No \\ 
        \cite{yu2021digital} & 2021 & Extension to COBie\cite{cobieontology} and OntoProg\cite{ontoprogontology} ontologies & COBie and OntoProg ontologies & Public Infrastructure & Physical, DT, Organizational & No \\ 
        \cite{farsi2021digital} & 2021 & A DT ontology for predicting lifecycle cost estimation in manufacturing & - & Manufacturing & DT, Application & No \\ 
        \cite{massel2021digital} & 2021 & An ontology for solar power plants to be used in DTs & - & Energy & Physical & No \\ 
        \cite{fujii2021digital} & 2021 & Smart home DT ontology & - & Smart Home & Physical & No \\ 
        \cite{zhao2021cutting} & 2021 & CNC Machining Ontology & - & Manufacturing & Physical, DT & No \\ 
        \cite{lu2021cognitive} & 2021 & A DT ontology for complexity management & - & Digital Twin & DT, Organizational & Yes \\ 
        \cite{zheng2021knowledge} & 2021 & Mechanical products ontology & - & Manufacturing & Physical, DT & No \\ 
        \cite{grebenyuk2021technological} & 2021 & Tech infrastructure management ontology & - & IT & DT, Organizational & No \\ 
        \cite{bamunuarachchi2021framework} & 2021 & A DT ontology extending author's earlier work\cite{bamunuarachchi2020cyber}, SOSA and SSN\cite{ssn-sosa} ontologies & Author's previous ontology, SOSA and SSN ontologies & Manufacturing & Physical, Communication, DT & No \\ 
        \cite{chukkapalli2021cyber} & 2021 & - & Smart farming ontology\cite{chukkapalli2020ontologies} & Smart Farming & DT & Yes \\ 
        \cite{saratha2021digital} & 2021 & - & Uses BFO\cite{bfoontology} as the top-level ontology and then 5 more ontologies that are specific to the use case described, see Table 2 in the paper. & Digital Twin & Physical, DT & No \\ 
        \cite{amar2021knowledge} & 2021 & DT fault management ontology & - & Manufacturing & Physical, DT & No \\ 
        \cite{khiat2020towards} & 2021 & Inflammatory bowel disease ontology & Based on 3 existing medical ontologies\cite{humandiseaseontology}\cite{fmaontology}\cite{medicineontology} & Medicine & Physical, Application & Yes \\ 
        \cite{chukkapalli2021ontology} & 2021 & An ontology for fisheries & Platys ontology\cite{zavala2015platys} & Smart Fisheries & Physical & No \\ 
        \cite{mavrokapnidis2021linked} & 2021 & - & Brick\cite{brickontology}, BOT\cite{botontology}, PROPS\cite{propsontology} and BEO\cite{beoontology} ontologies & Building Management & Physical, DT & Yes \\ 
        \cite{xu2021digital} & 2021 & Industrial robot control ontology & - & Manufacturing & Physical, DT, Application & No \\ 
        \cite{rovzanec2103actionable} & 2021 & Production planning and demand forecasting ontology & BFO\cite{bfoontology} & Manufacturing & DT, Application, Organizational & Yes \\ 
        \cite{Li2021CosimulationOC} & 2021 & An ontology for co-simulation of complex engineered systems & - & Digital Twin & Physical, DT & No \\ 
        \cite{d2021top} & 2022 & Top-level DT ontology & BFO\cite{bfoontology} & Digital Twin & Communication, DT & No \\ 
        \cite{akroyd_harper_soutar_farazi_bhave_mosbach_kraft_2022} & 2022 & OntoLandUse, OntoCropMapGML and OntoCropEnergy & - & Digital Twin & Physical & Yes \\ 
        \cite{SAHLAB2022463} & 2022 & Top-level ontology of mechatronic systems & - & Manufacturing & DT & No \\ 
        \cite{data7080105} & 2022 & Cultural heritage ontology & - & Cultural & Physical & No \\ 
        \cite{munker2021online} & 2022 & Internet of Construction On-Site Ontology (IoC-OSO) & DOCK, MASON\cite{masonontology}, MaRCO\cite{marcoontology}, MSO(no citation found), and \cite{zeng2017scenario} & Construction & Physical, Organizational & No \\ 
        \cite{kalidindi2022digital} & 2022 & - & Uses an existing materials ontology\cite{voigt2021materials} & Materials Science & Physical, DT & Yes \\ 
        \cite{abadi2022smart} & 2022 & Digital Twin Manufacturing Ontology (DTM-Onto) & - & Manufacturing & Physical, DT, Organizational & No \\ 
        \cite{doi:10.1080/17538947.2022.2127949} & 2022 & Railway DT Ontology & - & Public Infrastructure & Physical & Yes \\ 
        \cite{farghaly2022construction} & 2022 & Construction Programme \& Production Control Ontology & - & Construction & DT, Organizational & No \\ 
        \cite{ariansyah2022enhancing} & 2022 & DT and non-DT system interoperability ontology & - & Business & DT, Organizational & No \\ 
        \cite{van2022toward} & 2022 & Tactile internet ontology for tactile devices & - & IT & Physical & No \\ 
        \cite{9868776} & 2022 & An extended version of RealEstateCore ontology\cite{hammar2019realestatecore} & RealEstateCore & Building Management & Physical & No \\ 
        \cite{skobelev2022development} & 2022 & SDTP crop ontology & Author's previous work \cite{skobelev2020development, skobelev2021development} & Smart Farming & Physical & No \\ 
        \cite{kanak2022bimyverse} & 2022 & - & BIM, GIS and IoT ontologies (no specific ontology is cited) & Building Management & Physical, DT & No \\ 
        \cite{poudel2022integrated} & 2022 & A top-level manufacturing ontology & - & Manufacturing & Physical, DT, Application & No \\ 
        \cite{karmakar2022sdpm} & 2022 & An IoT device ontology & - & IoT & Communication, DT & No \\ 
        \cite{fierro2022notes} & 2022 & - & Brick\cite{brickontology} & Building Management & Physical & No \\ 
        \cite{huang2022enabling} & 2022 & - & MarCO\cite{marcoontology} & Manufacturing & Organizational & No \\ 
        \cite{d2022detecting} & 2022 & - & BFO\cite{bfoontology}, Common Core Ontology and IoF-Core\cite{iofcoreontology} & Manufacturing & DT & Yes \\ 
        \cite{wang2022digital} & 2022 & Clamping system ontology & - & Manufacturing & Physical, DT & No \\ 
        \cite{zheng2022semantic} & 2022 & Aircraft assembly system, manufacturing requirements and architecture model ontologies & IoF-Core\cite{iofcoreontology}, BFO\cite{bfoontology} & Manufacturing & Physical, DT, Application & Yes \\ 
        \cite{zheng2022blockchain} & 2022 & Mechanical products and a DT state modification ontology & - & Manufacturing & Physical, DT & No \\ 
        \cite{khan2022platology} & 2022 & Construction DT ontology & - & Construction & Physical, DT & No \\ 
        \cite{katsigarakis2022digital} & 2022 & COGITO & BEO\cite{beoontology}, BOT\cite{botontology}, GEO\cite{geoontology} & Construction & Physical, DT, Application, Organizational & No \\ 
        \cite{ma2022digital} & 2022 & IoT device ontology & SSN\cite{ssn-sosa} & IoT & Physical, DT & Yes \\ 
        \cite{donkerscreating} & 2022 & Occupant Feedback Ontology & - & Building   Management & DT, Application & Yes \\ 
        \cite{santos2022towards} & 2022 & O3POntology & BFO\cite{bfoontology}, GeoCore\cite{geocoreontology}, and IoF-Core\cite{iofcoreontology} & Energy & Physical, DT & No \\ 
        \cite{li2022cognitive} & 2022 & System of Systems (DTs) ontology & BFO\cite{bfoontology}, IoF specification\cite{iofspecs} & Digital Twin & Physical, DT & No \\ 
        \cite{jia2023simple} & 2023 & - & Machine tool ontology\cite{jia2022simple} & Manufacturing & Physical & Yes \\ 
        \cite{gowripeddi2023digital} & 2023 & Intrusion Detection System Ontology & - & IT Security & DT & No \\ 
        \cite{niccolucci2023heritage} & 2023 & Heritage DT ontology & - & Cultural & Physical, DT & No \\ 
        \cite{jiang2023intelligent} & 2023 & Building Fire Protection Ontology & - & Building Management & Physical, DT & No \\ 
        \cite{arsiwala2023digital} & 2023 & - & RealEstateCore Ontology\cite{hammar2019realestatecore} & Building Management & Physical, DT & Yes \\ 
        \cite{xie2023digital} & 2023 & - & Brick\cite{brickontology} & Building Management & Physical & No \\ 
\end{longtblr}
\twocolumn

\appendix

 \bibliographystyle{elsarticle-num} 
 \bibliography{cas-refs}





\end{document}